\pdfoutput=1
\documentclass[11pt]{article}

\usepackage{ACL2023}
\usepackage{graphicx}
\usepackage{multirow}
\usepackage{caption}
\usepackage{float}
\usepackage{tabularx, makecell}
\usepackage{times}
\usepackage{booktabs} 
\usepackage{siunitx}
\usepackage{lipsum}
\usepackage{latexsym}
\usepackage{array}
\usepackage{amsmath}
\usepackage{subcaption}

\usepackage[T1]{fontenc}
\usepackage[utf8]{inputenc}

\usepackage{microtype}

\usepackage{inconsolata}

\title{Large Language Models are Visual Reasoning Coordinators \\
Can VLMs be used on videos for action recognition?}

\author{Harsh Lunia \\
  IIIT Hyderabad \\ Mathematics of Generative Modeling, Dr. Pawan Kumar \\
  \texttt{harsh.research@iiit.ac.in} }

\begin{document}
\maketitle
\begin{abstract}

Recent advancements have introduced multiple vision-language models (VLMs) demonstrating impressive commonsense reasoning across various domains. Despite their individual capabilities, the potential of synergizing these complementary VLMs remains underexplored. The Cola Framework addresses this by showcasing how a large language model (LLM) can efficiently coordinate multiple VLMs through natural language communication, leveraging their distinct strengths. We have verified this claim on the challenging A-OKVQA dataset, confirming the effectiveness of such coordination.

Building on this, our study investigates whether the same methodology can be applied to surveillance videos for action recognition. Specifically, we explore if leveraging the combined knowledge base of VLMs and LLM can effectively deduce actions from a video when presented with only a few selectively important frames and minimal temporal information. Our experiments demonstrate that LLM, when coordinating different VLMs, can successfully recognize patterns and deduce actions in various scenarios despite the weak temporal signals. However, our findings suggest that to enhance this approach as a viable alternative solution, integrating a stronger temporal signal and exposing the models to slightly more frames would be beneficial.

\end{abstract}

\section{Introduction}

Visual reasoning is a crucial task that requires models to not only comprehend and interpret visual information but also apply high-level cognition to derive logical solutions \cite{johnson2017clevr, malkinski2023review, zakari2022vqa}. Effective visual reasoning necessitates a model to possess both strong visual perception and robust logical reasoning abilities. Traditional visual reasoning models often depend on complex architectures \cite{yi2018neural, mao2019neuro} and struggle to generalize beyond their training datasets \cite{zellers2019recognition, park2020visualcomet}. 

In contrast, large language models (LLMs) have shown impressive commonsense reasoning capabilities in natural language processing (NLP) applications, even in zero-shot settings \cite{brown2020language}. Socratic Models \cite{zeng2022socratic} further enhance these capabilities by facilitating communication between VLMs and LLMs through prompt engineering, unlocking zero-shot multimodal reasoning. This integration demonstrates the potential for advanced visual reasoning by combining the strengths of both VLMs and LLMs.

Interested in exploring how to integrate homogeneous expert models (e.g., multiple VLMs) with a large language model (LLM) in a coordinative framework, \cite{chen2024large} introduces Cola, a novel ensemble approach that employs an LLM as a coordinator among multiple VLMs. Traditional model ensemble methods typically focus on adjusting model weights \cite{ilharco2022patching} or combining predictions \cite{wortsman2022model}, which can be cumbersome to implement with widely-used end-to-end black box model APIs, such as GPT-4, Google Bard, and Anthropic Claude.

Recent research on augmented LLMs, such as \cite{schick2024toolformer}, has focused on enabling LLMs to utilize external tools comprehensively. However, the potential of prompt ensembles to aggregate outputs from multiple models has not been fully explored. In contrast, \cite{chen2024large} demonstrated that Cola leverages language prompts generated by multiple expert models to effectively create model ensembles, offering a new paradigm in coordinating VLMs through an LLM.

\textbf{Human Action Recognition Using Coordinated VLM Outputs: }
Human activity recognition (HAR) is a challenging task that involves detecting human actions in complex interactions without vocal communication. Human movements range from simple arm or leg motions to intricate, coordinated movements involving the entire body. In video, these actions are represented by a sequence of frames that can be understood by analyzing the information across multiple frames in order. However, processing every frame in a video for accurate action prediction is inefficient due to the large number of frames. This highlights the need for methods that can predict actions using inherent knowledge of the physical world and keyframes from the video.

As mentioned earlier, each vision-language model (VLM) has its own set of complex patterns in its outputs. The Cola framework, which coordinates different VLMs, can leverage their knowledge bases to accurately predict human actions using only a few select important frames, potentially achieving high levels of accuracy.

Our contributions are as follows:

\begin{itemize}
    \item We evaluate and confirm the claim that an LLM can act as a coordinator for multiple VLMs, producing better results than traditional ensemble approaches based on combining predictions or adjusting model weights.
    \item We utilize the existing Cola framework and demonstrate its effectiveness on video data by applying it to a video understanding task, specifically human activity recognition.
    \item We present the effectiveness of the framework as a high-potential, intelligent alternative by testing it in a challenging problem setting. In this setting, we provide either no or weak temporal information and a selective few important frames (a maximum of 10) from the video, and still achieve noteworthy performance in action recognition.
\end{itemize}

\section{Methodology}
\subsection{Cola Framework}
\begin{figure}
    \centering
    \includegraphics[width=\linewidth]{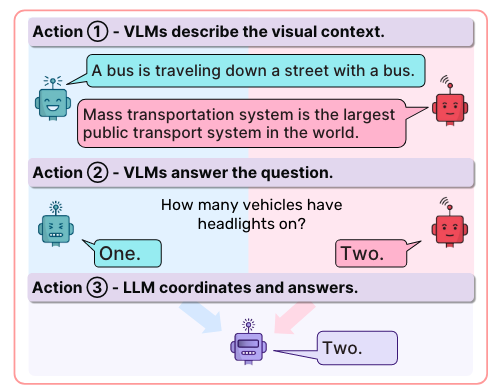}
    \caption{Cola coordinates multiple pretrained VLMs based on the visual context and plausible answers they provide.}
    \label{fig:cola_overview}
\end{figure}

\begin{figure}
    \centering
    \includegraphics[width=\linewidth]{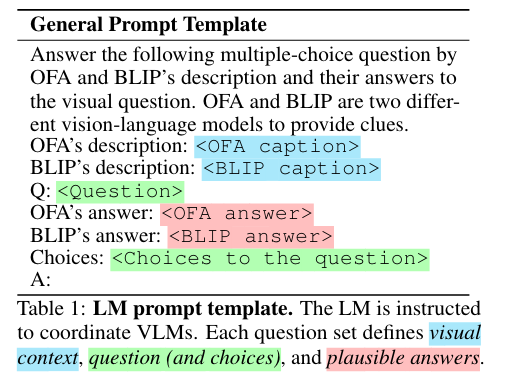}
    \caption{LM prompt template. The LM is instructed to coordinate VLMs. Each question set defines visual context, question (and choices), and plausible answers.}
    \label{fig:cola_vqa_template}
\end{figure}

An overview of the Cola framework is depicted in Figure \ref{fig:cola_overview}. We utilize OFA-Large \cite{wang2022ofa} and BLIP \cite{li2022blip} as the vision-language models (VLMs). The large language model (LLM) employed is FLAN-T5-Base \cite{chung2024scaling}, which features an encoder-decoder transformer architecture. For the Visual Question Answering (VQA) task, each VLM independently generates output captions and plausible answers. These outputs are then concatenated with the instruction prompt, the question with choices, and the captions to fuse all contexts, enabling the LLM to reason, coordinate, and provide an answer. The template used to verify the claim by \cite{chen2024large} is illustrated in Figure \ref{fig:cola_vqa_template}.

\subsection{Human Action Recognition}
To extend the study of using LLM-coordinated VLMs, we applied the Cola framework to surveillance videos.

We utilized the Surveillance Perspective Human Action Recognition (SPHAR) dataset \cite{sphar-dataset}, which is designed to analyze human activities in public places captured using cameras placed in typical surveillance settings. The videos in this dataset are sourced from multiple locations, making it diverse and representative of general, realistic scenarios.

Each video in the SPHAR dataset is associated with only one action at a time, making it suitable for a multi-class classification problem. The dataset covers a total of 14 distinct action classes.

The following steps were involved:

\subsubsection{Key Frame Selection}
Since VLMs operate on images rather than continuous frames or video sequences, it was necessary to extract keyframes that provide the most accurate and compact summary of the video content. We used the "Katna" library \cite{katna-library} for this purpose. A maximum of 10 keyframes were extracted per video. Some shorter videos (approximately 6 seconds) might have only one keyframe, resulting in no temporal information. For all videos, the extracted frames were not in chronological order, providing only weak temporal information for the VLMs. This increased the difficulty for the VLM/LLM coordinated framework, testing their knowledge base and strategizing capabilities for any noteworthy performance.

\begin{figure}
    \centering
    \includegraphics[width=\linewidth]{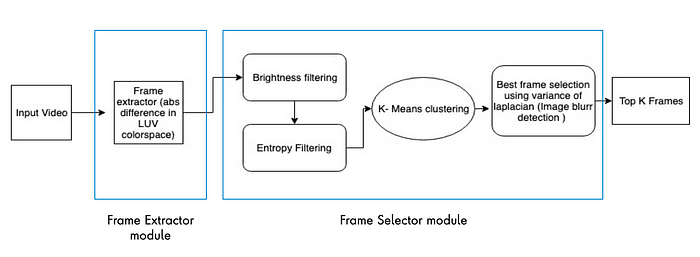}
    \caption{High-level architecture of the keyframe selection.}
    \label{fig:high_level_architecture}
\end{figure}

The selection process involved a combination of image processing-based filtering and classical statistical methods:
\begin{itemize}
    \item Frames that are sufficiently different from previous ones were initially selected using absolute differences in the LUV color space.
    \item These frames were further filtered based on brightness and entropy scores.
    \item K-Means clustering was applied to the frames using image histograms.
    \item The best frame from each cluster was selected based on the variance of the Laplacian (image blur detection).
\end{itemize}

\subsubsection{VLM Querying}
Once the keyframes were extracted, each VLM was queried for every keyframe from every video with the question: 
\begin{quote}
    “What action is happening in the frame?”
\end{quote}

\subsubsection{LLM Training}
\begin{figure}
    \centering
    \includegraphics[width=\linewidth]{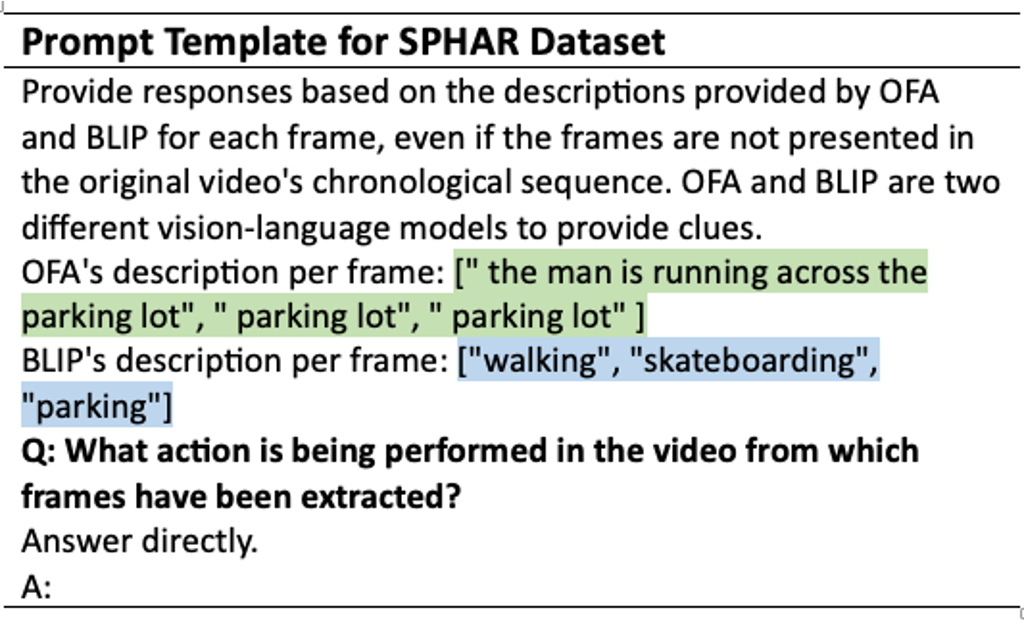}
    \caption{Custom template construction: Outputs from the queried VLMs for each keyframe are collated and used to create a template for LLM training against the correct target action name.}
    \label{fig:custom_template}
\end{figure}

Once all the VLMs were queried for each extracted keyframe, their outputs were collated to construct a custom template, as illustrated in the Figure \ref{fig:custom_template}. This template was then used to train the LLM against the correct anticipated target action name.

\section{Results}
\subsection{Validating LLM as Effective Coordinators: Core Claim Verification}
\begin{figure}[H]
    \centering
    \begin{subfigure}[b]{0.45\columnwidth}
        \centering
        \includegraphics[width=\textwidth]{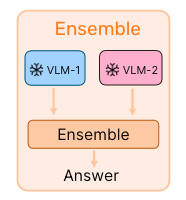}
        \label{fig:ensemble_paradigm}
        \caption{}
    \end{subfigure}
    \hfill
    \begin{subfigure}[b]{0.45\columnwidth}
        \centering
        \includegraphics[width=\textwidth]{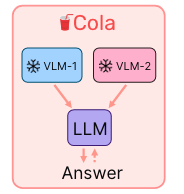}
        \label{fig:cola_paradigm}
        \caption{}
    \end{subfigure}
    \caption{Comparison between Paradigms}
    \label{fig:figures_side_by_side}
\end{figure}
\textbf{Ensemble Modeling} can be considered the most basic baseline for aggregating VLMs. It represents the base performance that the combination of VLMs can achieve on the target task when not trained. It aggregates multiple models’ predictions in order to improve the overall performance (Figure \ref{fig:figures_side_by_side}a). It's implemented as an averaging ensemble of cosine similarity between VLM output and each choice of a question as our ensemble baseline.
$$
P (v, q) = \frac{1}{n}\sum_{i=1}^{n} P_i(v, q)
$$
\textbf{Cola-FT}, Instruction Tuning of Cola is initialized with pretrained checkpoints. Given the question q based on the image v, the LM predicts the answer in the form of sequence  
$$
s_{v,q} = LLM(Prompt(v, q))
$$
Only the LLM is fine-tuned (not the VLMs) to adhere to the common paradigm of ensemble modeling and simplify the method (Figure \ref{fig:figures_side_by_side}b). Due to resource constraints, we verified the claim by \cite{chen2024large} using a smaller LLM model with 248 million parameters. While this reduced the accuracy, as shown in Table \ref{tab:reproduced_results}, it confirms that the LLM can effectively coordinate different VLMs to produce better results than a traditional ensemble approach. This further reinforces the paper's argument by demonstrating that even a significantly smaller LLM (almost 45 times smaller) can validate the core claim.

\begin{table}[ht]
\scriptsize 
\caption{Results on the A-OKVQA dataset comparing different methods. The original COLA-FT (Finetuned) method reported in the paper outperforms the baseline ensemble. Our reproduced version, using a smaller LLM model, shows slightly reduced performance compared to the original COLA-FT but still achieves higher accuracy than the baseline ensemble.}
\label{tab:reproduced_results}
\renewcommand\tabularxcolumn[1]{m{#1}}
\renewcommand\arraystretch{1.2}
\begin{tabularx}{\linewidth}{@{} >{\bfseries}X X X S[table-format=2.2] @{}}
    \toprule
\thead{Method} & \thead{ VLM \\Models} & \thead{LLM\\Model} & {\thead{Accuracy (\%)}}   \\
\midrule
\makecell{Ensemble \\ (Baseline) }      & \multirow{3}{*}{\makecell{BLIP (384M) \\ + \\ OFA (472 M) }} & \makecell{\centering -} & 56.6\\
\makecell{Cola-FT \\ (Paper)} &  & \makecell{FLAN-T5-xxl \\ (11B)} & 77.7\\
\makecell{Cola-FT \\ (Reproduced)}      &  & \makecell{FLAN-T5-base \\ (248 M) } & 59.48\\
\bottomrule
\end{tabularx}
\end{table}

\subsection{Effective Utilization of Different VLMs via LLM Coordination for Video-based Human Action Recognition}

\begin{table}[ht]
\scriptsize 
\caption{
Performance comparison of different VLMs coordinated by an LLM on a video dataset for Human Action Recognition. The result is compared with a state-of-the-art (SOTA) model that utilizes strong temporal information and processes more than twice the number of frames (24). Highest performing results are highlighted in bold. }
\label{tab:sphar_results}
\renewcommand\tabularxcolumn[1]{m{#1}}
\renewcommand\arraystretch{1.2}
\begin{tabularx}{\linewidth}{@{} >{\bfseries}X X X S[table-format=2.2] @{}}
    \toprule
\thead{Model} & \thead{Recall} & \thead{Precision} & {\thead{F1 score}}   \\
\midrule
\makecell{Ours}      & \makecell[r]{0.5941} & \makecell[r]{0.5608} & 0.5660 \\
\makecell{\textbf{2D-CNN+SPBD LSTM} \\ \cite{manaf2021novel}} & \makecell[r]{\textbf{0.9090}} & \makecell[r]{0. \textbf{9757}} & \textbf{0.9064} \\
\bottomrule
\end{tabularx}
\end{table}

The table presented in Table \ref{tab:sphar_results} showcases the outcomes of applying the Cola framework to carefully selected keyframes extracted from surveillance videos for human action recognition, juxtaposed with a state-of-the-art (SOTA) result from a separate study \cite{manaf2021novel}. The observed disparity in performance can be attributed to the SOTA model's incorporation of robust temporal information, enabling sequential processing of frames, and handling more than double the number of frames (24) compared to the COLA approach. Despite operating with significantly fewer frames and under the constraint of unordered frame sequences, our method demonstrates a commendable grasp of the task, yielding respectable performance.

\subsubsection{Analysis and Improvement Scope}
\begin{figure}
    \centering
    \includegraphics[width=\linewidth]{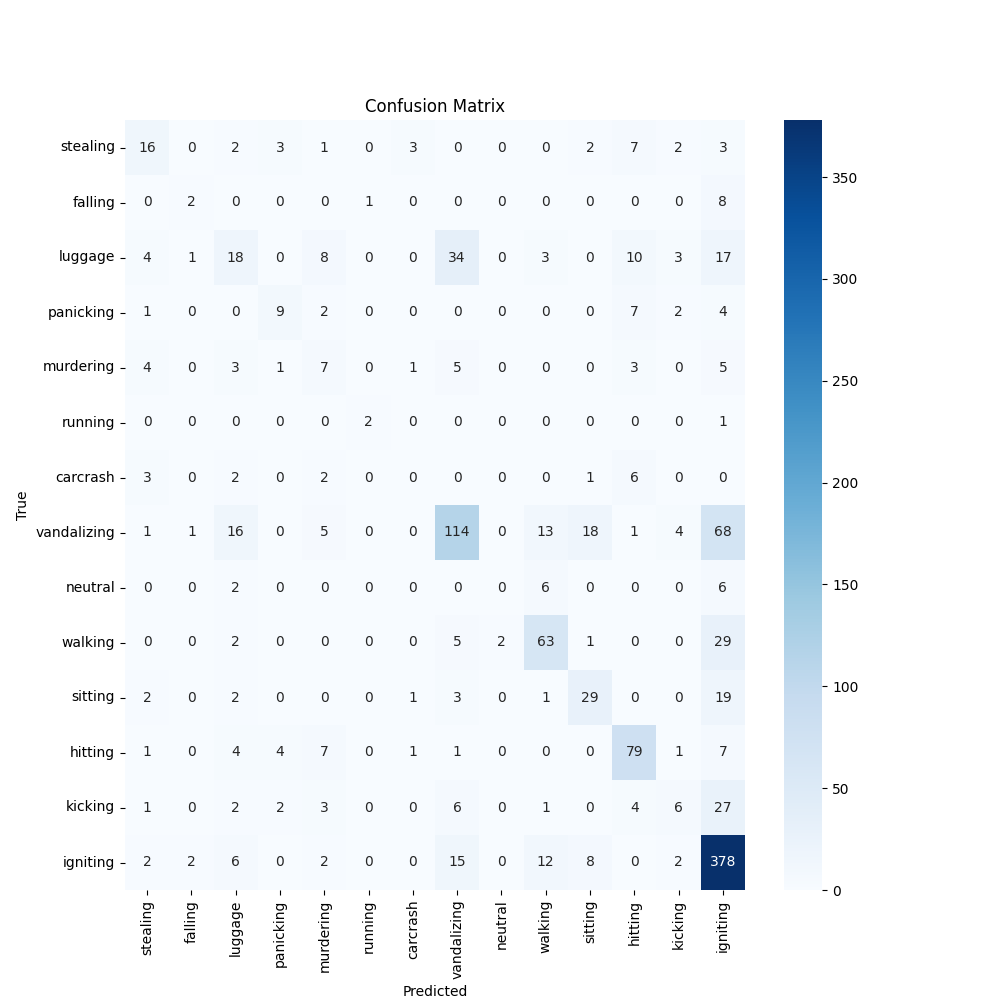}
    \caption{Confusion matrix illustrating the classification performance of our approach for human action recognition on the surveillance video dataset. Rows represent the actual classes, while columns represent the predicted classes. Higher values along the diagonal indicate accurate classifications, while off-diagonal values indicate miss-classifications.}
    \label{fig:confusion_matrix}
\end{figure}

The confusion matrix demonstrates our approach's performance across various action categories, excelling in identifying actions like stealing, vandalizing, walking, sitting, hitting, and igniting with robust accuracy. However, moderate performance is noted for actions such as panicking, murdering, and running, indicating areas for improvement. Challenges arise in accurately detecting anomalies within the car-crash category, highlighting the difficulty in distinguishing between anomalies and regular patterns during keyframe extraction. Moreover, the complexity of the SPHAR dataset, despite its diversity, presents additional hurdles as some samples contain actions that are not mutually exclusive. For example, neutral videos may be misinterpreted as hitting due to high-five actions, while luggage videos may be mistaken for walking as they involve humans carrying luggage. These complexities underscore the need for robust algorithms capable of discerning subtle differences and handling overlapping action categories effectively. We believe that enhancing submodules and utilizing larger LLMs/VLMs can leverage the deep understanding gained by these models to address these nuances and achieve improved performance.

Also, the task is made more challenging by the limited temporal knowledge of frames in the input template. Additionally, scaling the number of frames based on video length, restricted to just 10 frames, may not offer the optimal solution, particularly when the keyframe extractor fails to identify keyframes without any false negatives. Furthermore, the reliance on image processing and statistical methods often causes the keyframe extractor to overlook genuine action frames. Improving the accuracy of the extractor, while minimizing false negatives, is crucial for enhancing overall performance.

\section*{Acknowledgement}
I express gratitude to Prof. Pawan Kumar for proposing this paper as a study and analysis project within the Mathematics in Generative Modeling course. Through discussions with him, we explored the potential extension of the paper to video analysis, leading to valuable insights and laying the groundwork for further exploration and potential future research endeavors. Additionally, I extend my appreciation to the authors of the paper \cite{chen2024large} for introducing the straightforward concept of Cola and providing a comprehensive analysis of the framework. Their meticulous examination of various visual understanding tasks using different VLMs and LLMs of varying sizes instilled confidence in the effectiveness of the approach. Moreover, while reviewing the paper, I was impressed by their exploration of diverse questions within this framework, such as scaling studies, explainability of answers, visualization of important tokens, and transferability of Cola across out-of-distribution tasks.

\bibliography{custom}
\bibliographystyle{acl_natbib}

\end{document}